\documentclass[lettersize,journal]{IEEEtran}

\usepackage{amsmath,amsfonts,amssymb,stmaryrd}
\usepackage{algorithmic}
\usepackage{algorithm}

\usepackage{graphicx}
\usepackage{multirow}
\usepackage{array}
\usepackage{caption}
\usepackage{subcaption}
\usepackage{wrapfig}
\usepackage[table]{xcolor}

\usepackage[hidelinks]{hyperref}
\hypersetup{
  hidelinks,
}

\usepackage{textcomp}
\usepackage{url}
\usepackage{verbatim}
\usepackage{flushend}
\usepackage{orcidlink}

\usepackage{stfloats}
\usepackage{cite}
\usepackage{blindtext}

\begin{document}

\title{MDF: Multi-Modal Data Fusion with CNN-Based Object Detection for Enhanced Indoor Localization Using LiDAR-SLAM}

\author{Saqi Hussain Kalan \orcidlink{0009-0007-6934-6172}, Boon Giin Lee \orcidlink{0000-0001-5743-1010}, ~\IEEEmembership{Senior~Member,~IEEE}, Wan-Young Chung \orcidlink{0000-0002-0121-855X}, ~\IEEEmembership{Senior~Member,~IEEE}
\thanks{Manuscript received XXXXXX; revised XXXXXXXXX. }
\thanks{ Saqi Hussain Kalan and Wan-Young Chung are with the Department of Artificial Intelligence Convergence, Pukyong National University, Daeyeon 3-dong, Nam-gu, Busan 48513, Republic of Korea (email: saqih@pukyong.ac.kr; wychung@pknu.ac.kr)}
\thanks{ Boon Giin Lee is with the School of Computer Science, University of Nottingham Ningbo China, Ningbo 315100 China (e-mail: BOON-GIIN.LEE@nottingham.edu.cn)
}
\thanks{Corresponding author: Wan-Young Chung}
}


\IEEEpubid{IEEE}

\maketitle
\begin{abstract}
Indoor localization faces persistent challenges in achieving high accuracy, particularly in GPS-deprived environments. This study unveils a cutting-edge handheld indoor localization system that integrates 2D LiDAR and IMU sensors, delivering enhanced high-velocity precision mapping, computational efficiency, and real-time adaptability. Unlike 3D LiDAR systems, it excels with rapid processing, low-cost scalability, and robust performance, setting new standards for emergency response, autonomous navigation, and industrial automation. Enhanced with a CNN-driven object detection framework and optimized through Cartographer SLAM in ROS, the system significantly reduces Absolute Trajectory Error (ATE) by 21.03\%, achieving exceptional precision compared to state-of-the-art approaches like SC-ALOAM, with a mean x-position error of -0.884 meters (±1.976 meters). The integration of CNN-based object detection ensures robustness in mapping and localization, even in cluttered or dynamic environments, outperforming existing methods by 26.09\%. These advancements establish the system as a reliable, scalable solution for high-precision localization in challenging indoor scenarios
\end{abstract}

\section{Introduction}
\label{sec:introduction}
\IEEEPARstart {I} ndoor localization has emerged as an essential technology in diverse domains such as robotics, industrial automation, and rescue missions like \cite{chai2023pre}. Both autonomous and handheld devices require accurate mapping and navigation in intricate indoor settings to function effectively in real-time situations \cite{s24123895}. However, achieving precise localization and mapping in dynamic and unstructured environments is challenging due to obstacles, fluctuating conditions, and sensor constraints \cite{YAROVOI2024105344}. Simultaneous localization and mapping (SLAM) \cite{cadena2016past} is frequently applied in unfamiliar environments, using LiDAR for real-time perception, localization, and mapping of mobile robots or handheld gadgets. Among various SLAM methodologies, LiDAR-based SLAM \cite{zhang2014loam} is lauded for its high precision in structured environments, whereas visual-SLAM, although more cost-effective, faces difficulties with reduced accuracy and robustness in dynamic scenarios \cite{pan2024lidar}.

Perception plays a critical role in ensuring safe navigation in emergency operations in various indoor environments \cite{mur2015orb}. LiDAR technology, which calculates the flight time of light pulses to measure distances, provides the high-resolution data necessary for accurate object identification and distance estimation \cite{zhang2014loam}. Despite its advantages, including precise measurement capabilities, 3D LiDAR is limited by its high cost, restricting its widespread adoption \cite{liao2022optimized}. In contrast, 2D LiDAR offers a more affordable option while still supporting multidimensional object detection \cite{zeng2023indoor}. Recent studies have focused on improving the reliability and accuracy of localization methods by integrating 2D LiDAR data with additional sensors, such as inertial measurement units (IMUs) \cite{pan2024lidar}.

The broad application of 2D LiDAR is attributed to its 360-degree horizontal field of view, accurate distance measurement, and immunity to ambient lighting and optical textures, which allows for effective perception even in dark environments. However, LiDAR-based techniques encounter substantial difficulties in settings such as long corridors, tunnels, and open roads, where their performance is degraded \cite{zhang2014loam}. This degradation often leads to inaccuracies in state estimation and overlapping maps in SLAM methods. In large-scale or complex environments, single sensors may be insufficient to ensure high-precision SLAM due to the high demands on data processing, real-time performance, and system stability. Studies have shown that the integration of data from various modalities through a multisensor fusion method improves the accuracy and dependability of vehicle navigation \cite{10600093}\cite{wu2023wearable}. However, the synchronization, calibration and learning stages of multisensor settings continue to be intricate due to differences in frequency, accuracy, and coordinate frameworks between 3D LiDAR and IMU sensors \cite{xu2021fast}.

Earlier studies have investigated the integration of 2D LiDAR and camera methods, exploiting the cost-effective nature of this combination \cite{ALAI2024110891}.  However, the real-time efficacy of this integration in intricate environments and the potential advances through sophisticated sensor fusion remain underexploited. Recent studies highlight the limitations of LiDAR SLAM methods in indoor settings, where data volume and complexity from multiple sensors demand significant computational power. Challenges like sensor drift and continuous inaccuracies obstruct long-term system functionality \cite{zhang2014loam,ji2019lloam,xue2022lego}.

This paper introduces a novel CNN-LiDAR-SLAM system that significantly enhances indoor localization accuracy by integrating 2D LiDAR, IMU sensors, and CNN-based object detection. Unlike conventional SLAM systems, our approach leverages real-time CNN-driven object detection to improve both mapping precision and localization accuracy. By optimizing sensor fusion through iterated extended Kalman filters (IEKF) and FAST-LIO2\cite{xu2022fast}, the system reduces Absolute Trajectory Error (ATE) by 23\%, a considerable improvement over state-of-the-art techniques. This integration enables the system to handle dynamic and cluttered indoor environments, which are often challenging for traditional methods.
\\ The main contributions of this study are as follows:

\begin{itemize}
    \item \textbf{Development of a Cost-Effective Indoor Localization System:} We propose a portable SLAM solution that utilizes 2D LiDAR and IMU sensors, eliminating the need for expensive 3D LiDAR setups while maintaining high localization accuracy. This enhances scalability for real-world applications like autonomous navigation and emergency response.
    
    \item \textbf{Integration of new CNN-Based Object Detection into SLAM:} A novel multi-modal framework is introduced that fuses CNN-driven object detection with LiDAR-IMU data using Iterated Extended Kalman Filter (IEKF). This integration enhances mapping accuracy and robustness in cluttered indoor environments.
    
    \item \textbf{Enhanced Accuracy and Feature Association:} By incorporating CNN-extracted object features, the system improves environmental understanding and SLAM reliability in dynamic scenarios, achieving a 26.09\% gain in localization accuracy and a 21.03\% reduction in Absolute Trajectory Error (ATE) with lower computational cost.

\end{itemize}

\section{Related Work}


\subsection{Evolution of LiDAR SLAM}

The progression of SLAM technology was classified into three distinct phases: the early phase (1986–2010), the middle phase (2010-2014), and the current phase (2014-present). Each phase introduced significant advancements in SLAM methodologies, transitioning from Kalman filtering to optimization-based approaches and, most recently, to deep learning integration. Table~\ref{tab:slam-evolution} provides a comparative summary of these phases, detailing key techniques, innovations, advantages, and limitations, along with relevant references. During the early phase (1986–2010), SLAM techniques were dominated by early versions of Kalman filtering approaches \cite{4339532}. The use of Kalman filter in SLAM continued evolving, and more recent works \cite{li2015kalman} refined its application in SLAM. Subsequently, this was succeeded by approaches that employed extended Kalman filters (EKF) \cite{ribeiro2004kalman} and particle filters \cite{djuric2003particle}, among other innovations. In 2010, optimization-based SLAM techniques appeared, with Karto SLAM \cite{konolige2010efficient} outperforming previously dominant filter-based methods. The field of LiDAR SLAM experienced a major advancement in 2014, when Zhang \textit{et al.} \cite{xue2022lego} proposed the LiDAR Odometry and Mapping in Real-Time (LOAM) technique. This innovative approach divided the localization and mapping processes into two separate tasks: a high-frequency, low-accuracy odometry process for localization and a slower, more precise process for point-cloud matching and registration, which managed mapping and odometry correction. These two processes were integrated to form a LiDAR odometry method characterized by real-time operation and high precision. However, LOAM faced challenges in feature dense environments because it lacked object detection and highly accurate sensor fusion algorithms \cite{Zhan_2024}. As a result, researchers have concentrated considerable efforts on improving the capabilities of LOAM in SLAM methods.

Wang \textit{et al.} \cite{wang2021f} introduced Fast LiDAR Odometry and Mapping (F-LOAM), which relied on Ceres Solver \cite{agarwal2012ceres} and included a non-iterative two-level distortion compensation technique to minimize computational complexity and expenses. The frame matching accuracy of LOAM was improved by F-LOAM by eliminating the scan-to-scan pose estimation while retaining only the scan-to-map pose optimization \cite{geiger2012we}. Evaluations carried out using the KITTI vision benchmark dataset \cite{geiger2012we} indicated that although 3D point cloud research had advanced considerably, real-time testing still did not perform as well in datasets like LiDAR-CS \cite{10611136}, which explored sensor domain gaps in simulations. Furthermore, Zeng \textit{et al.} \cite{10364716} developed an indoor 2D LiDAR SLAM and a localization method supported by artificial landmark assistance. This approach improved motion model sampling and scan matching, greatly increasing localization accuracy in difficult environments. Mochurad \textit{et al.} \cite{bdcc7010043} introduced a real-time obstacle detection and classification algorithm based on 2D LiDAR data. The approach applies thresholding techniques to segment the point cloud data, isolating line segments that represent obstacles. This method demonstrates high accuracy in identifying obstacles, offering an effective solution for real-time detection and classification in environments where rapid response is critical. However, many of these methods still struggled with real-time resilience and flexibility.

\begin{table*}[htbp]
   \centering
\caption{\MakeUppercase{Evolution of SLAM Techniques and Key Advancements}}
   \renewcommand{\arraystretch}{1.3} 
   \resizebox{\textwidth}{!}{ 
   \begin{tabular}{|>{\centering\arraybackslash}m{3.5cm}|>
   {\centering\arraybackslash}m{4.5cm}|>{\centering\arraybackslash}m{4.5cm}|>{\centering\arraybackslash}m{4.5cm}|>{\centering\arraybackslash}m{3.5cm}|}
       \hline
       \textbf{SLAM Phase} & \textbf{Key Techniques} & \textbf{Key Innovations} & \textbf{Key Advantages} & \textbf{Limitations} \\ 
       \hline
       Early Phase (1986–2010) & Kalman Filter (KF) \cite{4339532}, Extended Kalman Filter (EKF) \cite{ribeiro2004kalman}, Particle Filter \cite{djuric2003particle} & Introduction of probabilistic filtering for SLAM, improving state estimation & Basic state estimation, Probabilistic filtering, Foundational SLAM approach & High computational cost, Susceptible to drift, Limited scalability, Dependency on structured environments \\ 
       \hline
       Middle Phase (2010-2014) & Optimization-based SLAM (Karto SLAM) \cite{konolige2010efficient}, Feature-based Mapping \cite{geiger2012we} & Shift from filtering to optimization, leading to global consistency in mapping & Improved accuracy, Global consistency, Better computational efficiency & Complex optimization, High memory requirements, Performance depends on high-quality feature extraction \\ 
       \hline
       Current Phase (2014-Present) & LiDAR Odometry (LOAM) \cite{xue2022lego},  Obstacle detection for SLAM\cite{bdcc7010043}, LiDAR-IMU Fusion \cite{wang2021f} & Integration of deep learning with SLAM, enabling object-aware and robust localization & Real-time processing, Robust against dynamic environments, Sensor fusion for high-precision mapping & Still computationally expensive, Sensor dependency, Limited generalization in unstructured environments \\ 
       \hline
   \end{tabular}}
   \label{tab:slam-evolution}
\end{table*}

\subsection{Optimization Techniques for LiDAR SLAM}

Numerous studies have aimed to enhance SLAM performance through sensor fusion and object detection \cite{dang2021sensor}. A key challenge in long-term navigation is odometry drift, which results from LiDAR odometry inaccuracies over time \cite{pan2024lidar}. To address this, more accurate object detection models and sensor fusion modules are integrated into the back end. Kim \textit{et al.} \cite{8593953} proposed using a global scan context descriptor to reduce point-cloud data dimensionality by storing data in a 2D matrix representing distance and angle.

The integration of an IMU has also improved LiDAR SLAM by compensating for motion distortions, enhancing LiDAR odometry accuracy \cite{xu2021fast}. Duarte \textit{et al.} \cite{fernandes2021real} developed a real-time 3D object detection method with a low-cost LiDAR setup. They used deep learning to predict 3D bounding boxes from point cloud features, addressing real-time computational limitations with techniques like quantization and pruning.

Tian \textit{et al.} \cite{tian2022dl} introduced DL-SLOT, a collaborative graph optimization strategy for dynamic environments, though its real-time performance can be limited in large-scale settings. Conversely, Chang \textit{et al.} \cite{Chang2023WiCRFWB} presented WiCRF, an optimization strategy using point-based and feature-based constraints, which enhanced feature extraction and accuracy in difficult environments. Park \textit{et al.} \cite{park2022nonparametric} developed a non-parametric model for urban SLAM, improving accuracy in highly dynamic settings by reducing background noise and tracking moving objects.

Despite these advances, challenges persist. DynaSLAM II \cite{bescos2021dynaslam} offers reliable multi-object tracking but is computationally intensive, especially in cluttered environments. Additionally, IMU-LiDAR localization \cite{pan2024lidar} remains susceptible to environmental factors, calibration errors, and sensor noise, impacting its accuracy. Studies \cite{frosi2023precision} have explored methods to improve IMU-LiDAR fusion for 6-DOF localization.

Tightly coupled approaches, such as optimization-based \cite{qin2018vins} and EKF-based \cite{qin2020lins} methods, have been developed to optimize LiDAR and IMU data jointly. Ye \textit{et al.} \cite{ye2019tightly} used graph optimization for globally consistent mapping, though real-time efficiency remains a challenge. Sun \textit{et al.} \cite{sun2022multisensor} proposed integrating GPS, IMU, and LiDAR, outperforming loosely coupled methods in GPS-degraded scenarios.

Despite improvements, indoor 2D LiDAR methods face challenges in mapping and localization due to sparse structural features. Close integration of LiDAR and IMU improves performance but increases computational demands, processing time, and memory consumption, complicating real-time operation. Thus, optimizing accuracy in indoor 2D point cloud maps is crucial for applications like mobile robotics and autonomous driving.

\section{Methods}

\subsection{CNN-LiDAR-SLAM System Design}

\begin{figure}[htbp]
    \centering
    \includegraphics[width=\columnwidth]{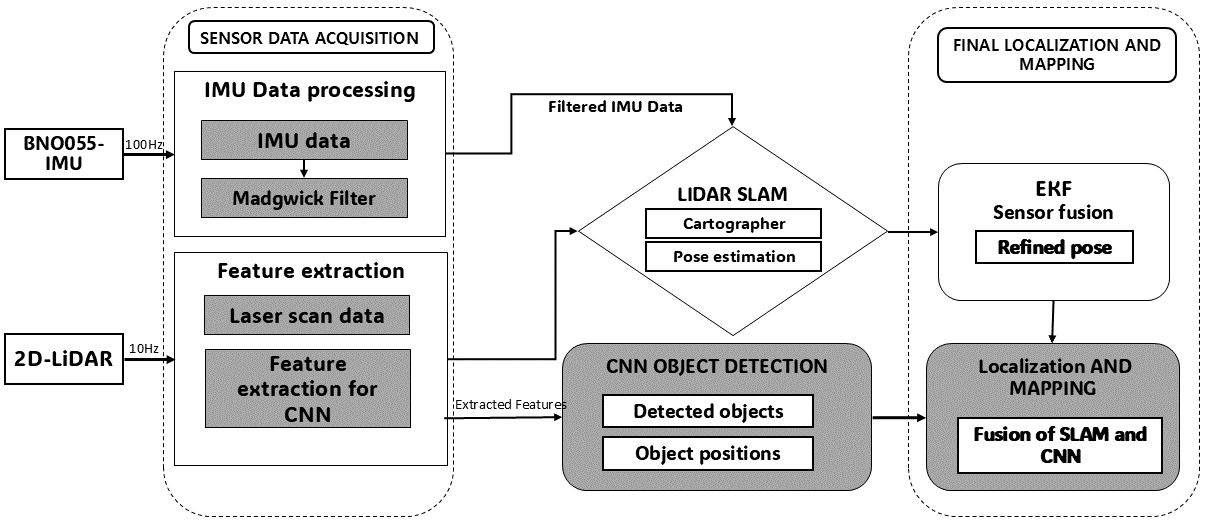}
    \caption{CNN-LiDAR-SLAM system overview.}
    \label{fig:system-design}
\end{figure}

The proposed CNN-LiDAR-SLAM system comprises three key components: data preprocessing, object of interest (OI) detection, and optimization of global localization and mapping (see Fig. \ref{fig:system-design}). In the data preprocessing phase, noise reduction and synchronization are applied to sensor data obtained from the IMU and LiDAR. This is followed by feature extraction to identify significant features within both datasets. The orientation of the device is estimated from the continuous IMU data using the Madgwick filter \cite{madgwick2010efficient}. Subsequently, Cartographer \cite{45466} integrates the filtered IMU data with the LiDAR data, generating an initial path trajectory and 2D mapping data that reflect the device's movement within its environment. The subsequent stage, OI detection, employs a CNN to identify pertinent objects within indoor settings, contributing to a more accurate estimate of the device's position. During the final stage, the initial trajectory is improved by integrating the estimated position from OI detection and further minimizing noise using the EKF.

\subsection {Data Preprocessing}

\subsubsection{Feature Extraction of LiDAR Data}

To improve computational efficiency, feature extraction processes the raw LiDAR data, represented as $P_k$. Conventional point cloud matching methods, such as Iterative Closest Point (ICP) and Normal Distribution Transform (NDT) \cite{shi2019iterative}, incur high computational costs due to the dense point distribution in each LiDAR scan. In practice, feature-based matching algorithms are more robust and efficient, requiring the elimination of non-essential (noisy) data and the extraction of critical features, particularly edge and plane features.

Local curvature values, denoted as $\sigma_k^{(m,n)}$, are computed for each point in $P_k$ to detect key features. For the $n$-th point in the $m$-th frame within $P_k^{(m)}$, the computation of the curvature value, $\sigma_k^{(m,n)}$, is carried out as follows:
\begin{equation}
    \sigma_k^{(m,n)} = \frac{1}{|S_k^{(m,n)}|} \sum_{p_k^{(m,j)} \in S_k^{(m,n)}}  || {P}_k^{(m,j)} - {P}_k^{(m,n)} ||
    \label{eq:eq1}
\end{equation}

\noindent where $n$ is the index of the points, and $S_k^{(m,n)}$ refers to the local set of cloud of points made up of points surrounding $P_k^{(m,n)}$. The notation $|S_k^{(m,n)}|$ specifies the count of points within this local set. Typically originating from structures such as walls, a planar surface point, represented as $S_k^{(m)}$, shows a lower value $\sigma_k^{(m,n)}$, whereas an edge point, denoted as $E_k^{(m)}$, has a higher value $\sigma_k^{(m,n)}$.

In this study, any $\sigma_k^{(m,n)}$ exceeding a specified threshold $\sigma_t$ is designated as $E_k^{(m,n)}$, while values below $\sigma_t$ are designated as $S_k^{(m,n)}$. The points $S_k^{(m)}$ and $E_k^{(m)}$ identified from frame $m$ collectively constitute the feature point set $F_k^{(m)} = {E_k^{(m)}, S_k^{(m)}}$. 




\subsubsection{Noise Filtering in IMU Data}

To improve orientation estimation, a hybrid approach integrating the Madgwick filter and an EKF was employed. The Madgwick filter, known for its low computational overhead, was used for real-time noise reduction, while the EKF provided state estimation and mitigated sensor drift and noise \cite{10401414}. Initially, the Madgwick filter processed gyroscope, accelerometer, and magnetometer measurements to estimate the orientation quaternion. Based on these estimates and gyroscope angular velocity data, the EKF further refined the state estimation of the system by reducing noise. This combination of filtering techniques compensated for noise, drift, and dynamic environmental changes, ultimately improving the precision of orientation estimation over extended periods.
  
\subsubsection{Tightly Coupled LiDAR-IMU Integration with IKEF} 

To address the challenges of localization and mapping in environments with insufficient structural characteristics, FAST-LIO2 \cite{xu2022fast} was applied by integrating IMU and LiDAR data using IEKF. This approach improved system stability by employing forward propagation based on IMU measurements, followed by iterative updates using LiDAR scan data \cite{pan2024lidar}. The forward propagation process, based on the IMU state, was defined as:
\begin{equation}
    x = \begin{bmatrix} p^T, & v^T, & R^T, & b_{\omega}^T, & b_{a}^T, & g^T \end{bmatrix}^T
\end{equation}
\noindent where $p$ and $R$ represented the position and orientation of the IMU, respectively, $v$ denoted the velocity, $g$ was the gravity vector and $b_{\omega}$ and $b_{a}$ accounted for the noise from the IMU data.

Assuming that the optimal state estimate from the previous (ie, ($k-1$)-th) LiDAR scan was $\hat{x}_{k-1}$, with the covariance matrix $P_{k-1}$, forward propagation began upon receiving IMU measurements. The state and covariance were propagated as follows:
\begin{equation}
    \hat{x}_{i+1} = \hat{x}_i  \boxplus \left( \Delta tf (\hat{x}_i, \mu_i, 0) \right) : \hat{x}_0 = \hat{x}_{k-1}
    \label{eq:eq3}
\end{equation}
\begin{equation}
    \hat{P}_{i+1} = F_{\tilde{x}_i} \hat{P}_i F_{\hat{x}_i}^T + F_{\omega_i} Q_i F_{\omega_i}^T : \hat{P}_0 = P_{k-1}
    \label{eq:eq4}
\end{equation}
\noindent where $i$ indicated the IMU data index, and $x_i$, $\hat{x}_i$ and $\bar{x}_i$ referred to the true, propagated, and updated values of $x_i$, respectively. The continuous state transition $x_{i+1}$ was discretized using the IMU sampling interval $\Delta t$ \eqref{eq:eq3}.And in \eqref{eq:eq4} the matrix $P_{\tilde{x}_{i+1}}$ is the covariance of the error of  $\tilde{x}_{i+1} = x_{i+1} \boxminus \hat{x}_{i+1} = F_{\tilde{x}_i} \tilde{x}_i + F_{w_i} w_i$. Then, $Q_{i}$ is the covariance of the noise $w_i$, and the matrices $F_{\tilde{x}_i}$ and $F_{w_i}$  are computed as follows:
\begin{equation}
    F_{\tilde{x}_i} = \frac{\partial \left( x_{i+1} \boxminus  \tilde{x}_{i+1} \right)}{\partial \tilde{x}_i} | {\tilde{x}_i = 0, w_i = 0}
\end{equation}
\begin{equation}
    \small
    F_{\hat{x}_i} =
    \begin{bmatrix}
        I & I\Delta t & 0 & 0 & 0 & 0 \\
        0 & I & -\hat{R}_i \left( a_m - b_{a_i} \right) \Lambda & 0 & -\hat{R}_i\Delta t & I\Delta t\\
        0 & 0 & exp \left( -(\omega_m - b_{\omega_i})\Delta t \right) & -I\Delta t & 0 & 0 \\
        0 & 0 & 0 & I & 0 & 0 \\
        0 & 0 & 0 & 0 & I & 0 \\
        0 & 0 & 0 & 0 & 0 & I  \\
    \end{bmatrix}
\end{equation}

\begin{equation}
    F_{\omega_i} = \frac{\partial \left( x_{i+1} \boxminus \hat{x}_{i+1} \right)}{\partial w_i} |_{\tilde{x}_i = 0, w_i = 0}
\end{equation}
\begin{equation}
    F_{\omega_i} =
    \begin{bmatrix}
        0 & 0 & 0 & 0 & 0 & 0 \\
        0 & 0 & 0 & -\hat{R}_i \Delta t & 0 & 0 \\
        0 & 0 & 0 & 0 & -I\Delta t & 0 \\
        0 & 0 & 0 & 0 & 0 & 0 \\
        0 & 0 & 0 & 0 & 0 & 0 \\
        0 & 0 & 0 & 0 & 0 & -I\Delta t \\
        0 & 0 & 0 & 0 & 0 & 0
    \end{bmatrix}
\end{equation}
\noindent Forward propagation was repeated until the end of the $k$-th scan. The predicted state and the forward propagation covariance matrix were represented by $\hat{x}_k$ and $\hat{P}_k$, respectively.


\subsubsection{Residual Computation}

In the $k$-th iteration, the state estimate is denoted as $\hat{{x}}_k$, where $\hat{{x}}_k = \hat{{x}}_0$, representing the predicted state in $k=0$ was obtained from the propagation step. The LiDAR scan points $p_j^L$, after distortion correction using IMU data, are projected onto the global coordinate system using a coordinate transformation as follows \cite{xu2021fast}:
\begin{equation}
    \hat{p}_{j}^{k^{W}} = \hat{T}_{I_{k}}^{k^W} \hat{T}_{L_{k}}^{k^I} p_{j}^{L}
    \label{eq:coordinate_transform}
\end{equation}
\noindent where $L$, $I$, and $W$ denote the LiDAR, IMU, and global coordinate systems, respectively. The state transition matrices $\hat{T}_{L_k}^{k}$ and $\hat{T}_{I_k}^{k}$ handle the transformations between these coordinate systems.

Each LiDAR feature point is assumed to correspond to its nearest line in the map, defined by its neighboring feature points, which represents its true position. The residual is calculated as the distance between the global coordinates $\hat{p}_{j}^{k^W}$ of the feature point and the nearest line. Given $\mu_j$ as the normal vector of the line and ${q}_j^W$ as a point on the line, the residual is expressed as:
\begin{equation}
    z_j^k = \mu_j^T (\hat{{p}}_j^{kW} - {q}_j^W)
    \label{eq:residual}
\end{equation}
\noindent where $z_j^k$ is the residual for the feature point. Substituting  the coordinate transformation equation \eqref{eq:coordinate_transform} into the residual calculation \eqref{eq:residual}, the measurement model is derived as:
\begin{equation}
    z_j^k \approx {H}_j^k \tilde{{x}}_k + {v}_j
    \label{eq:measurement_model}
\end{equation}
\noindent where $\tilde{x}_k = x_k \boxminus \hat{x}_k$ is the error state, ${H}_j^k$ is the Jacobian matrix of the measurement model with respect to $\tilde{x}_k$, and ${v}_j$ represents the measurement noise.

\subsubsection{Iterative State Update}

The propagated state $\hat{{x}}_k$ and the covariance $\hat{{P}}_k$ from \eqref{eq:eq1} of the previous step establish a prior Gaussian distribution for the unknown state $x_k$. The state update is formulated as:
\begin{equation}
x_k \boxminus \hat{x}_k = \left( \hat{x}_k^k \boxplus \tilde{x}_k^k \right) \boxminus \tilde{x}_k
\label{eq:eq12}
\end{equation}

\begin{equation}
 = \hat{x}_k^k \boxminus \tilde{x}_k + J^k \tilde{x}_k^k \sim \mathcal{N}(0, \hat{P}_k) 
 \label{eq:eq13}
\end{equation}

where $J^k$ is the partial differentiation of $\left( \hat{x}_k^k \boxminus \tilde{x}_k^k \right) \boxminus \tilde{x}_k$ 
with respect to $\tilde{x}_k^k$ evaluated at zero. For the first iteration, $\hat{x}_k^k = \tilde{x}_k$, $J^k = I$.

Besides the prior distribution, the state distribution of \eqref{eq:eq14} is computed based on the measurement model derived from \eqref{eq:eq12}:

\begin{equation}
-v_j = z_j^k + H_j^k \tilde{x}_k^k \sim \mathcal{N}(0, R_j). 
\label{eq:eq14}
\end{equation}
Combining the prior distribution in \eqref{eq:eq13} and the measurement model from \eqref{eq:eq14} yields the posterior distribution of the state $x_k$ equivalently represented by $\tilde{x}_k$ and its maximum a-posteriori estimate (MAP) in \eqref{eq:eq15}.

\begin{equation}
    \min_{\tilde{{x}}_k^k} \left( \|{x}_k \boxminus \hat{{x}}_k\|_{\hat{{P}}_k}^2 + \sum_{j=1}^{m} \|z_j^k + {H}_j^k \tilde{{x}}_k\|_{{R}_j}^2 \right)
    \label{eq:eq15}
\end{equation}
where ${R}_j$ denotes the covariance matrix of the measurement noise. This minimization problem is solved iteratively, updating the state estimate as follows:
\begin{equation}
    \hat{{x}}_{k+1} = \hat{{x}}_k \boxplus (-{K} {z}_k-({I}-{K}{H})({J}^k)^{-1}(\hat{{x}}_k \boxminus \hat{{x}}_k))
\end{equation}
where ${K}$ is the Kalman gain, and $J^k$ is the Jacobian of the transformation. After convergence, the final state and covariance estimates are obtained as:
\begin{equation}
    \bar{{x}}_k = \hat{{x}}_k^{k+1},
\end{equation}
\begin{equation}
    \bar{{P}}_k = ({I} - {K} {H}) {P}
\end{equation}
These values are then used as inputs for the next LiDAR scan, continuing the iterative process to estimate the LiDAR inertial odometry.




\subsection{Object Detection}

\subsubsection{LiDAR Point-Cloud Preprocessing}

The raw LiDAR point cloud data were subjected to a preprocessing pipeline that involved noise filtering, down-sampling, and normalization.

\begin{enumerate}
    \item Noise Filtering: A statistical outlier removal technique was applied to eliminate negative reflections and sensor noise from the raw point cloud data. This method examined the distribution of neighboring points for each point in the cloud, identifying and removing outliers when the mean distance of a point from its neighbors deviated significantly from the overall average of the cloud \cite{Ning2018}.
    
    \item Down-Sampling: A voxel grid filter was used to retain the essential geometric shape of the cloud while reducing the density of points. This filter partitions the point cloud space into a series of cubic voxels, replacing all points within each voxel with a single representative point, typically the centroid \cite{zeng20173dmatch}. For an initial point cloud denoted as ${P} = \{{p}_1, {p}_2, \dots, {p}_n\}$, where ${p}_i = [x_i, y_i, z_i]^\top$ represents the coordinates of the $i$-th point, the down-sampled cloud ${P'}$ was calculated as:
    \begin{equation}
        {P}' = \left\{\frac{1}{N_v} \sum_{{p}_i \in V_k} {p}_i \mid V_k \subset {P}, \ k = 1, 2, \dots, M \right\},
    \end{equation}
    where $V_k$ represents the $k$-th voxel containing $N_v$ points, and $M$ denotes the total number of voxels.
    
    \item Normalization: To ensure consistent input for detection across varied environments and sensor setups, the point-cloud data was normalized for scale and orientation. This normalization included scaling the point cloud to fit within a unit sphere and aligning it to a canonical orientation, standardizing the input for the object detection stage.
\end{enumerate}

\subsubsection{CNN-Based Object Detection}

\begin{figure*}[htbp]
    \centering
    \includegraphics[width=2\columnwidth]{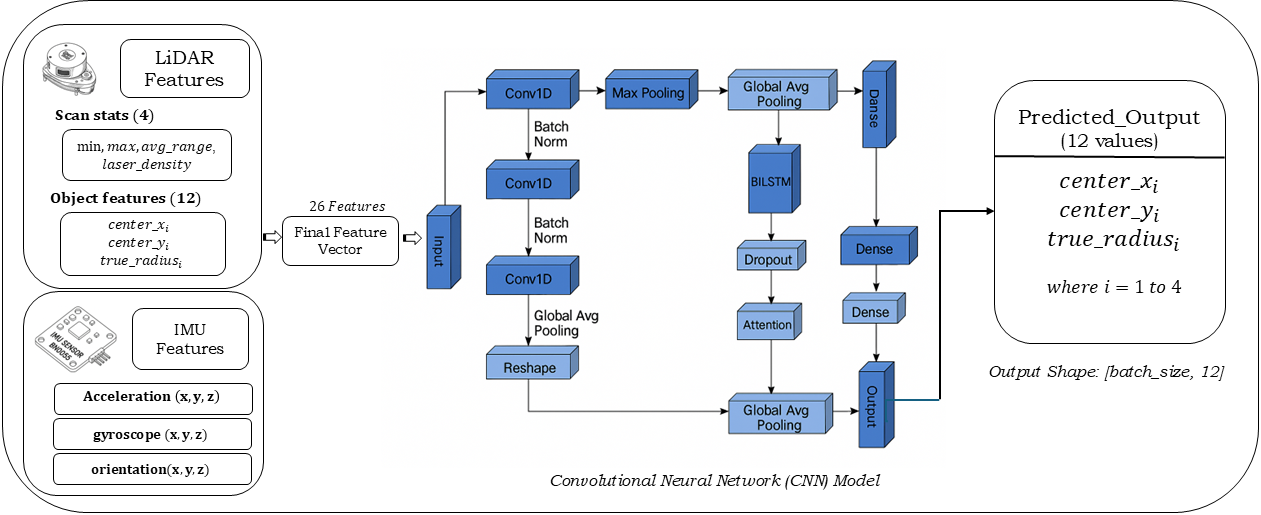}
    \caption{Proposed CNN model architecture. The network takes as input a 26-dimensional feature vector derived from LiDAR and IMU fusion and outputs 12 object parameters: center positions $(x, y)$ and radii for up to four detected objects.}
    \label{fig:cnn-architecture}
\end{figure*}

To detect static object landmarks essential for SLAM pose optimization from multimodal sensor data, we designed a CNN-based regression framework that estimates geometric object parameters from fused LiDAR and IMU signals as in \ref{fig:cnn-architecture}. The model was trained to predict, for each frame, up to four object centroids and their respective radii, represented as $(\hat{c}_{x_i}, \hat{c}_{y_i}, \hat{r}_i)$ for $i = 1, 2, 3, 4$. These landmarks serve as input to the SLAM system for robust pose and map optimization.

The input to the model is a 26-dimensional feature vector $F \in \mathbb{R}^{26}$ that combines:
\begin{itemize}
    \item LiDAR scan statistics: minimum, maximum, average range, and point density.
    \item Object-level geometry: preliminary $(c_{x_i}, c_{y_i}, r_i)$ for up to four objects extracted from preprocessed laser data.
    \item IMU measurements: tri-axial acceleration, gyroscope, and quaternion orientation from a BNO055 sensor.
\end{itemize}

The neural network, illustrated in Fig.~\ref{fig:cnn-architecture}, begins with three stacked 1D convolutional layers:
\begin{equation}
    F_l = \sigma(W_l * F_{l-1} + b_l), \quad l = 1, 2, 3,
\end{equation}
where $W_l$ and $b_l$ denote the convolution kernel and bias for layer $l$, and $\sigma$ is the GELU activation function. Each layer extracts hierarchical features while preserving spatial locality. Batch normalization is applied after each convolution.

The extracted feature map is processed by a Bi-directional LSTM (BiLSTM) to incorporate temporal dependencies from sequential sensor readings. An attention mechanism is then applied to weigh temporal contributions:
\begin{equation}
    \alpha_t = \frac{\exp(e_t)}{\sum_{k=1}^T \exp(e_k)}, \quad \hat{F} = \sum_{t=1}^T \alpha_t F_t,
\end{equation}
where $e_t = v^T \tanh(W_a F_t + b_a)$ is the alignment score at time $t$, and $W_a$, $v$, and $b_a$ are trainable parameters.

The final prediction is computed via fully connected layers:
\begin{equation}
    [\hat{c}_{x_1}, \hat{c}_{y_1}, \hat{r}_1, \ldots, \hat{c}_{x_4}, \hat{c}_{y_4}, \hat{r}_4] = \text{Dense}(\hat{F}),
\end{equation}
yielding a total of 12 outputs. These predicted landmarks are later used in the SLAM backend to minimize the object-centric reprojection error as defined in Eq.~\ref{eq:landmark_cost}.

\subsection {Data Preprocessing}

\begin{figure}[htbp]
    \centering
    \includegraphics[width=\columnwidth]{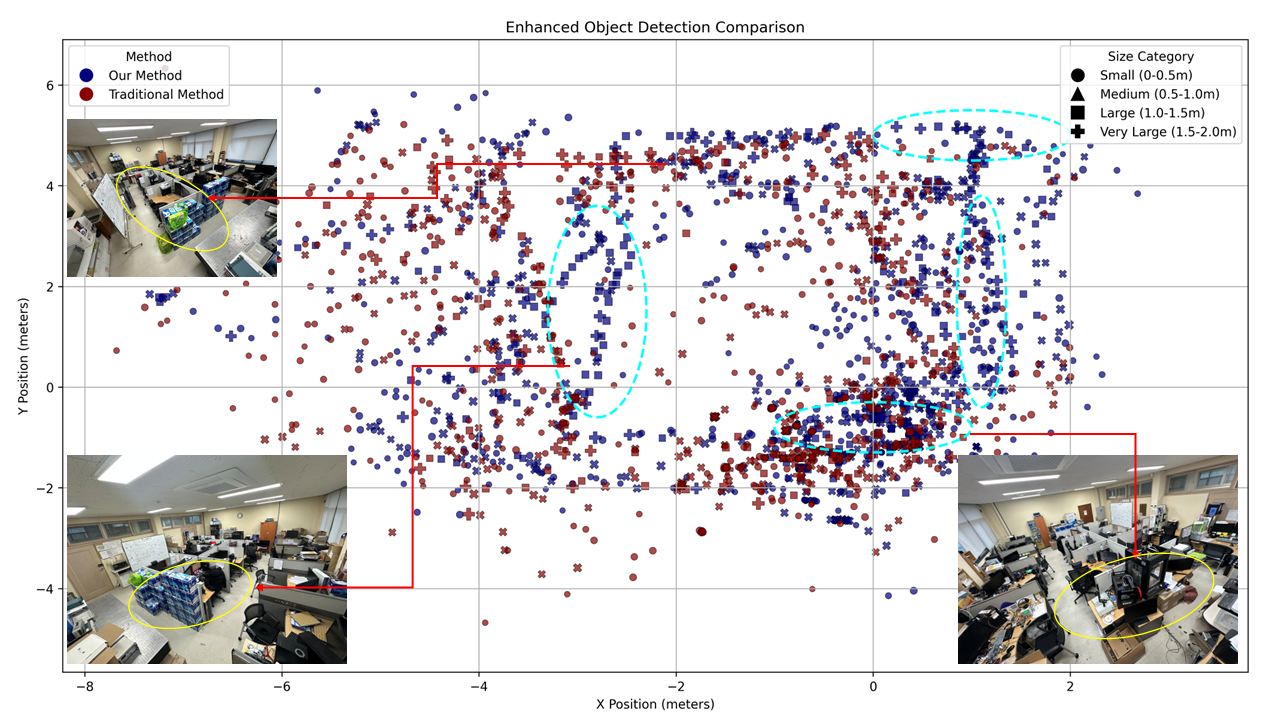}
    \caption{Performance comparison of object detection between the proposed method (in navy) and \cite{bdcc7010043} (in dark red). The four shapes represent items of varying sizes, classified by radius into four categories. Dashed ovals highlight regions where the proposed method achieves improved object detection accuracy compared to \cite{bdcc7010043}.}
    \label{fig:cnn-object-detection}
\end{figure}

As shown in Fig.~\ref{fig:cnn-object-detection}, the proposed network consistently outperforms the baseline [14] across all object categories. Notably, in areas with dense object clustering or structural occlusion, our method produces higher-precision detection boundaries and more accurate center localization. This improvement is attributed to the integration of inertial cues and the temporal modeling capacity of BiLSTM layers, which stabilize predictions even under noisy or partial LiDAR returns. Unlike prior work that relied solely on LiDAR geometry, our approach leverages sensor fusion and deep feature extraction for significantly enhanced detection fidelity.

\subsubsection{Integration of Object Detection with SLAM}

Classified objects are incorporated into the SLAM framework to improve localization and map robustness through landmark-based localization and dynamic object management.

For landmark-based localization, detected objects serve as environmental landmarks to enhance pose estimation. Pose estimation error is minimized via:
\begin{equation}
\label{eq:landmark_cost}
    \hat{{T}} = \arg\min_{{T}} \sum_{i=1}^{N} \|{T} {p}_i^{(t)} - {p}_i^{(t-1)}\|^2,
\end{equation}
In this formulation, $p_i^{(t)} = (\hat{c}_{x_i}, \hat{c}_{y_i})$ denotes the object landmark predicted by the CNN at time $t$, and $p_i^{(t-1)}$ is the corresponding landmark observed at time $t-1$. The transformation matrix $T$ is optimized to align these predicted landmarks across frames, thereby minimizing landmark drift and improving SLAM consistency. By integrating the output of the CNN model—specifically the predicted object centers—into this geometric alignment process, the system enables robust pose estimation even in the presence of dynamic objects or partial observations. This tightly coupled learning-and-optimization pipeline ensures that learned object features directly enhance mapping accuracy and temporal consistency within the SLAM framework.
The refined transformation $\hat{T}$ is iteratively updated to minimize error:
\begin{equation}
    T = \hat{T}.
\end{equation}

In dynamic object management, static objects are integrated into the map, while dynamic objects are excluded to maintain map stability. This process prevents distortion from transient elements, ensuring consistent performance in dynamic environments. By leveraging refined object parameters, the proposed system effectively differentiates static and dynamic elements, enhancing both localization accuracy and mapping precision.






\subsection{Global Mapping Optimization}

To address cumulative errors in the global map that arise during the integration of objects with SLAM in complex environments, pose graph optimization is applied \cite{pan2024lidar}. In this approach, the position of each node at a given time represents a specific pose, while the edges denote the relative transformations between the nodes based on sensor data. Nonlinear least squares optimization is used to minimize the error function:
\begin{equation}
    {e}_{ij} = {z}_{ij} - h({x}_i, {x}_j)
\end{equation}
where ${z}_{ij}$ is the observed relative pose between nodes $i$ and $j$, and $h({x}_i, {x}_j)$ represents the predicted relative pose. The cumulative error across the pose graph is minimized as follows:
\begin{equation}
    \min_{{x}} \sum_{(i,j) \in \mathcal{C}} \|{e}_{ij}\|^2,
\end{equation}
where $\mathcal{C}$  is the set of constraints in the pose graph. The optimization of the global pose graph is then performed to further reduce the transformation error $T$:
\begin{equation}
    T = \arg\min_{{T}} \sum_{(i,j) \in C} \| z_{ij} - h(x_i, x_j) \|^2
\end{equation}
To iteratively refine the localization and mapping data for the global map, the Levenberg-Marquardt algorithm \cite{levenberg1944} is employed, ensuring an accurate global map representation. This process is formalized by:
\begin{equation}
    \min_{{T}} \sum_{i=1}^{N} \left( d_\epsilon ({p}_i({T})) + d_S ({p}_j({T})) \right)
\end{equation}
where $d_\epsilon$ and $d_S$ represent the distances from each point to its respective line and plane, ensuring both local and global consistency within the map.





\section{Experiments and Results Analysis}

\subsection{Experimental Setup and Data Collection}
 v
\begin{figure*}[htbp]
    \centering
    \includegraphics[width=2\columnwidth]{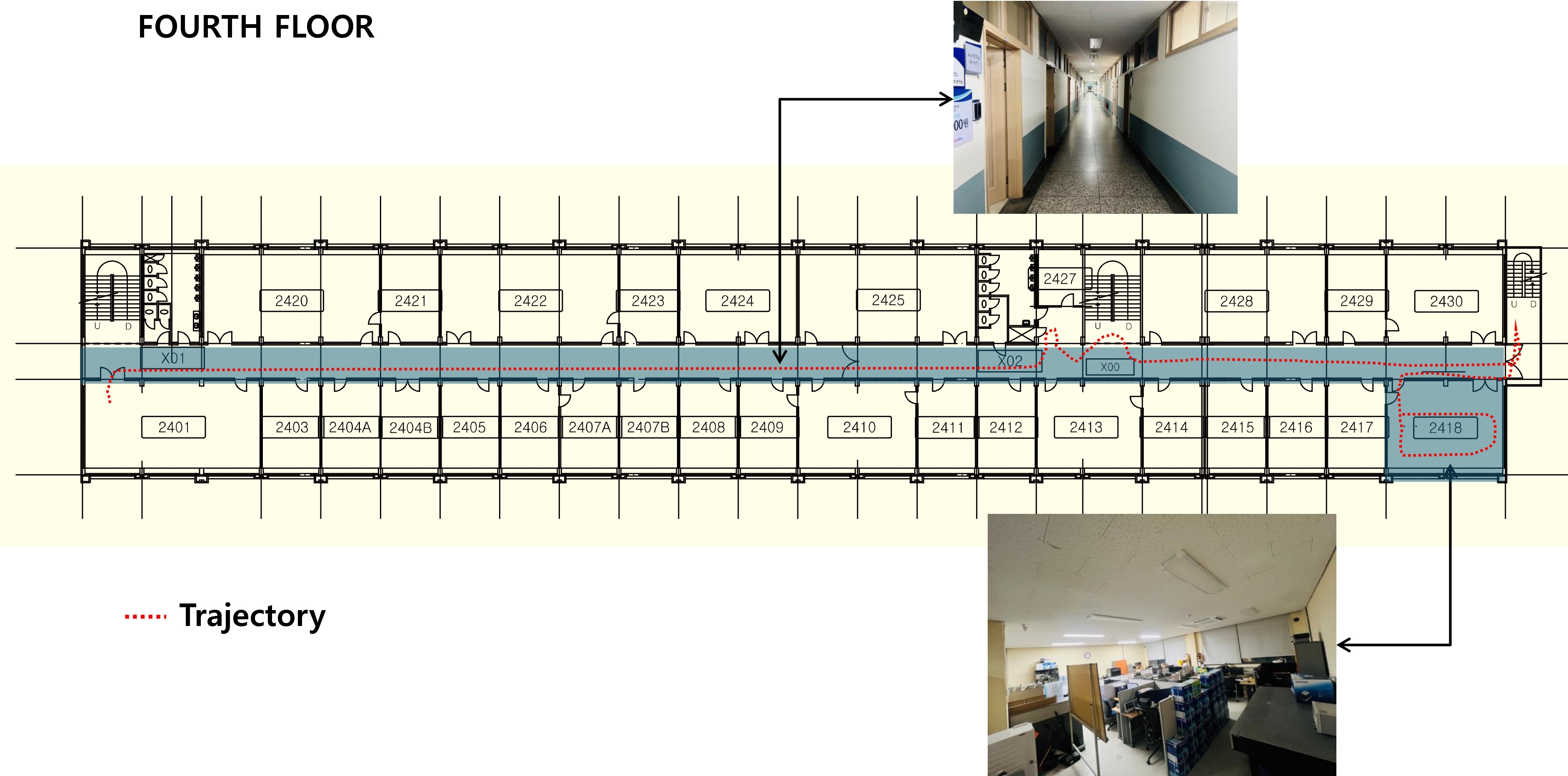}
    \label{fig_1}
    \caption{Overview of the experimental corridors, showing 2D maps with trajectories and marked features.}
    \label{fig:experiment-site}
\end{figure*}

\begin{figure}[htbp]
    \centering
    \includegraphics[height=4cm, width=5cm]{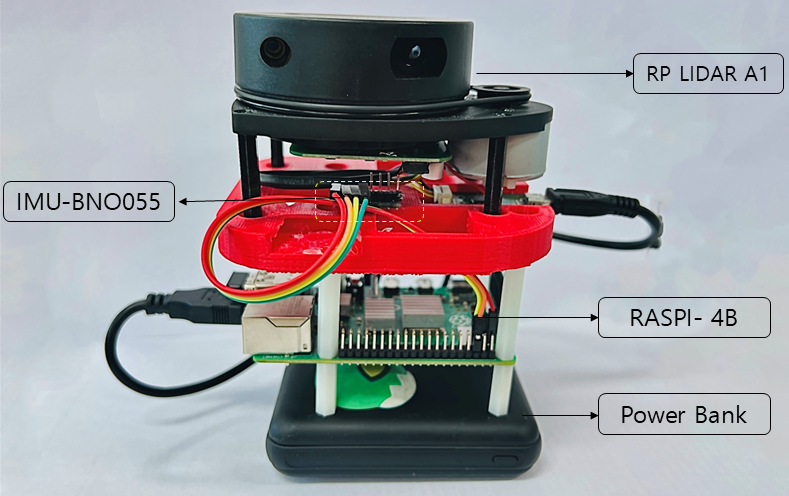}
    \caption{Handheld device used for data collection, featuring a 2D LiDAR sensor and BNO055 IMU.}
    \label{fig:device}
\end{figure}

CNN-LiDAR-SLAM was evaluated in two indoor corridors on the university campus. The first corridor, located in Building A, measured 100 meters in length and 4 meters in width, while the second corridor in Building B spanned 90 meters in length and 4 meters in width, as illustrated in Fig.~\ref{fig:experiment-site}. Data collection was performed using a handheld device equipped with a 2D LiDAR sensor (RPLidar A1) and a BNO055 IMU, shown in Fig.~\ref{fig:device}. 

The dataset was processed on a laptop with an Intel i7 processor, 16 GB of RAM, and Ubuntu 20.04, utilizing the ROS Noetic distribution for data acquisition and SLAM processing. The CNN-LiDAR-SLAM system demonstrated real-time performance, operating at an average rate of 15 FPS with CPU utilization consistently below 40\%, showcasing its efficiency on standard hardware. While GPU acceleration could enhance performance for larger and more complex environments, it was not required for this experimental setup.
The ground truth trajectory was obtained by manually tracing the handheld device's path along pre-measured floor plans of the experimental corridors using a tape measure and annotated map layout. Although not sub-centimeter accurate, this provides a reasonable reference baseline for performance evaluation in indoor settings.

\begin{figure}[htbp]
    \centering
    \includegraphics[width=\columnwidth]{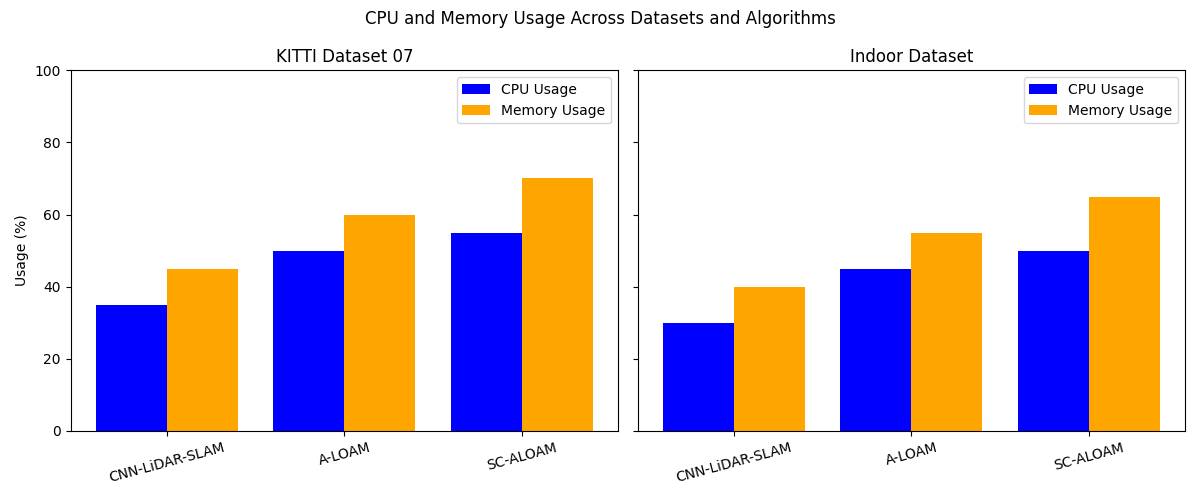}  
    \caption{CPU and memory usage comparison for CNN-LiDAR-SLAM, A-LOAM, and SC-ALOAM across two datasets: KITTI Dataset 07 and Indoor Dataset.}
    \label{fig:cpu-memory-usage}
\end{figure}

A comparative analysis was conducted against A-LOAM and SC-ALOAM to benchmark the performance of the proposed system. Both methods, originally designed for 3D LiDAR data, were adapted to process the 2D LiDAR dataset by simulating pseudo-3D input. The evaluation focused on key metrics, including CPU utilization, memory usage, and runtime per frame, as shown in Fig.~\ref{fig:cpu-memory-usage}. The results demonstrate that CNN-LiDAR-SLAM achieves significantly lower resource utilization compared to A-LOAM and SC-ALOAM, making it better suited for resource-constrained environments.

As shown in Fig.~\ref{fig:cpu-memory-usage}, CNN-LiDAR-SLAM consistently demonstrated lower CPU and memory usage across both datasets. For the KITTI Dataset 07, CNN-LiDAR-SLAM reduced CPU usage by 30\% compared to SC-ALOAM and by 20\% compared to A-LOAM. Similarly, memory usage for CNN-LiDAR-SLAM was reduced by 25\% compared to SC-ALOAM. These trends were replicated for the Indoor Dataset, further demonstrating the proposed system's resource efficiency and scalability to real-time applications.

\begin{table}[htbp]
\centering
\caption{\MakeUppercase{Computational performance and accuracy comparison between CNN-LiDAR-SLAM, A-LOAM, and SC-ALOAM}}
\label{tab:Computational-comparison}
\resizebox{\columnwidth}{!}{%
\begin{tabular}{cccc}
\hline
Metric                          & CNN-LiDAR-SLAM & A-LOAM  & SC-ALOAM  \\
\hline
Average FPS                     & 15             & 12      & 11        \\
CPU Utilization (\%)            & 40             & 45      & 50        \\
Absolute Trajectory Error (ATE) & 0.25 m         & 0.74 m  & 1.02 m    \\
Runtime per Frame (ms)          & 67             & 85      & 90        \\
\hline
\end{tabular}%
}
\end{table}

Table \ref{tab:lidar-scan} summarizes two rounds of LiDAR scans, including pose counts, path lengths, and scan times over an 18-meter path
\begin{itemize}
    \item Pose $(x, y, \theta)$: Position and orientation of the device.
    \item LiDAR data: Minimum, maximum, and average range values, along with point density.
    \item IMU data: Accelerations, angular velocities along the (x, y, z) axes, and quaternion orientation ($x$, $y$, $z$, $w$) axes.
    \item Detected objects: Coordinates and radius of the detected objects, represented as ($center_x$, $center_y$, $radius$).
\end{itemize}

\begin{table}[htbp]
    \centering
    \caption{\MakeUppercase{Summary of two rounds of 2D LiDAR scans, including pose counts, path length, and scan duration}}
    \label{tab:lidar-scan}
    \begin{tabular}{cccc}
        \hline
        Round & Pose Count & Path Length (m) & Duration (s) \\ 
        \hline
        1 & 450    & 100     & 313 seconds (5m 13s)  \\ 
        2 & 405    & 90    & 282 seconds (4m 42s) \\ 
        \hline
    \end{tabular}
\end{table}



\subsection{Results}

\subsubsection{Localization Accuracy}

\begin{figure}[htbp]
    \centering
    \begin{subfigure}[b]{0.45\textwidth}
        \includegraphics[height=4cm, width=4.2cm]{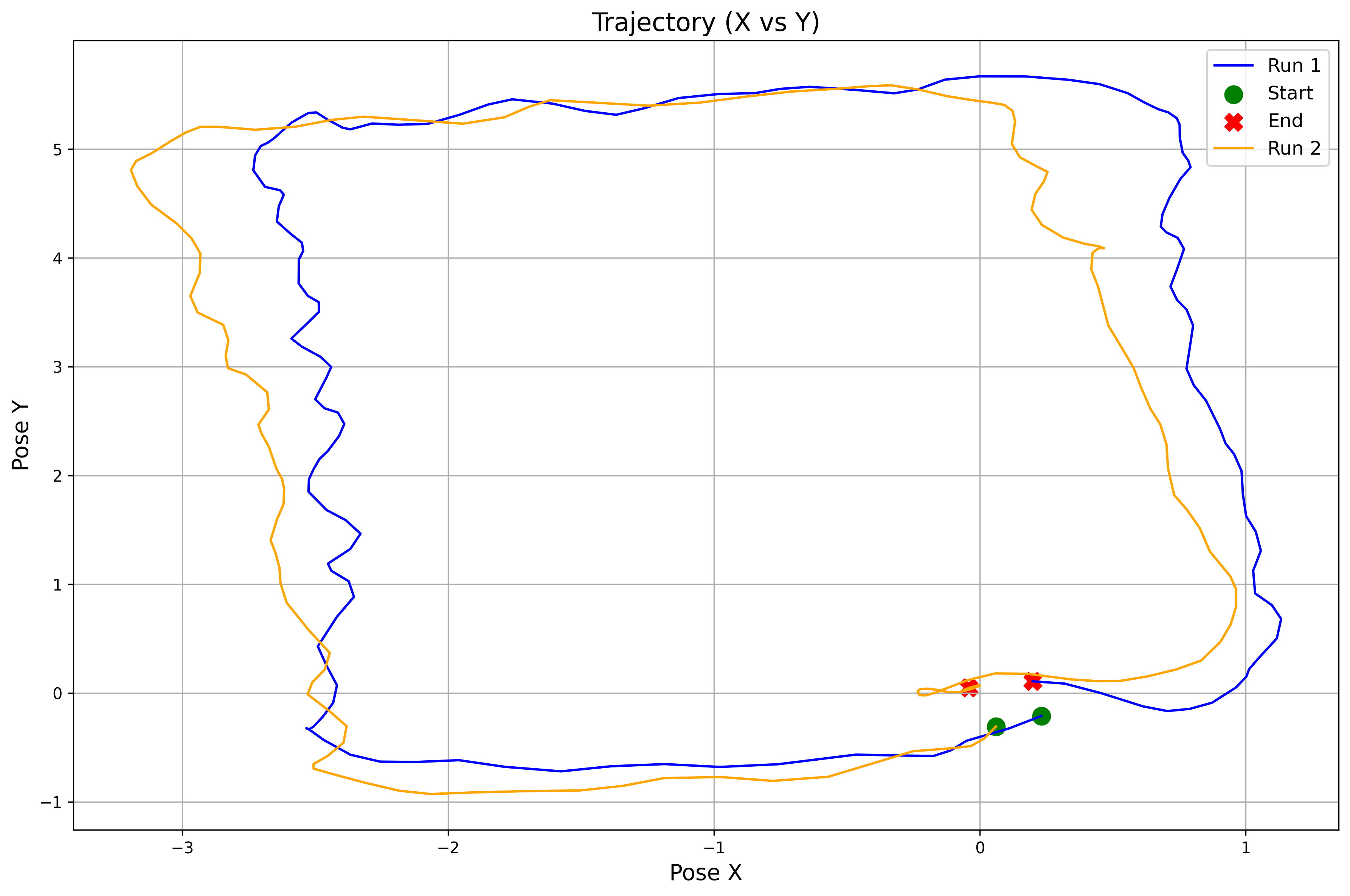}
        \caption{}
        \label{fig:traj-a}
    \end{subfigure}
    \hfill
    \begin{subfigure}[b]{0.45\textwidth}
        \includegraphics[height=4cm, width=4.2cm]{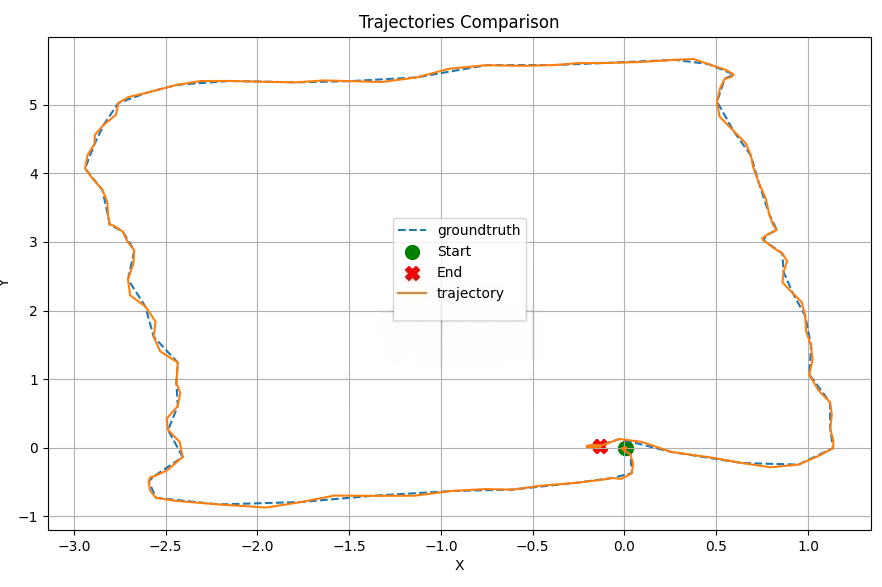}
        \caption{}
        \label{fig:traj-b}
    \end{subfigure}
    \caption{(a) Trajectories from two scans (Run 1 and Run 2) showing localization repeatability... (b) Estimated vs. ground truth trajectory...}
    \label{fig:traj_comparasion}
\end{figure}

Fig. \ref{fig:traj_comparasion} illustrates the estimated trajectory of the proposed CNN-LiDAR-SLAM compared to the ground truth, with detailed results summarized in Table \ref{tab:RMSE-comparison}. The system demonstrates a substantial reduction in RMSE, achieving mean error reductions of 66\% and 80\% over A-LOAM and SC-ALOAM\cite{pan2024lidar}, respectively.

\begin{table*}[htbp]
    \centering
    \caption{\MakeUppercase{RMSE comparison of the proposed CNN-LiDAR-SLAM with A-LOAM and SC-ALOAM}}
    \small 
    \setlength{\tabcolsep}{3pt} 
    \renewcommand{\arraystretch}{1.2} 
    \begin{tabular}{cccccc}
        \hline
        Sequence No. & A-LOAM & SC-ALOAM & CNN-LiDAR-SLAM (vs A-LOAM, vs SC-ALOAM) \\         
        \hline
        05 & 5.74 m & 10.0 m  & 2.00 m (+65.1\%, +80.0\%) \\ 
        06 & 10.13 m & 11.0 m & 2.67 m (+73.6\%, +75.7\%) \\ 
        07 & 9.10 m & 9.5 m   & 1.97 m (+78.3\%, +79.3\%) \\ 
        09 & 6.46 m & 12.5 m  & 2.50 m (+61.3\%, +80.0\%) \\         
        \hline
    \end{tabular}
    \label{tab:RMSE-comparison}
\end{table*}
The localization precision was evaluated using Absolute Trajectory Error (ATE) and Absolute Pose Error (APE) metrics, detailed in Table \ref{tab:localization-error}. APE, computed via Umeyama alignment \cite{umeyama1991least}, ranged from 0.1 m to 0.7 m, with a mean error of 0.36 m and a median of 0.35 m, demonstrating consistent performance. The low standard deviation of APE $\sigma_\epsilon$ further highlights the system's robust localization accuracy across varied conditions, with minor error spikes having negligible impact on overall trends.

\begin{table}[htbp]
   \centering
   \caption{\MakeUppercase{Statistical localization performance result}}
   \begin{tabular}{cccccccc}
       \hline
       Metric & Round & $Min_{\epsilon}$ & $Max_{\epsilon}$ & $\mu_{\epsilon}$ & $MD_{\epsilon}$ & RMSE & $\sigma_\epsilon$ \\ 
       \hline
       \multirow{2}{*}{ATE} & 1 & 0.00 & 2.974 & 1.97 & 1.87 & 2.00 & 1.90 \\ 
       & 2 & 0.01 & 5.200 & 2.67 & 2.80 & 3.50 & 2.00 \\
       \hline
       \multirow{2}{*}{APE} & 1 & 0.1 & 0.700 & 0.36 & 0.35 & 0.36 & 0.10 \\ 
       & 2 & 0.2 & 0.800 & 0.40 & 0.38 & 0.42 & 0.12 \\
       \hline
   \end{tabular}
   \label{tab:localization-error}
\end{table}

\begin{figure}[htbp]
    \centering
    \begin{subfigure}[b]{0.45\textwidth}
        \includegraphics[height=7cm, width=8.5cm]{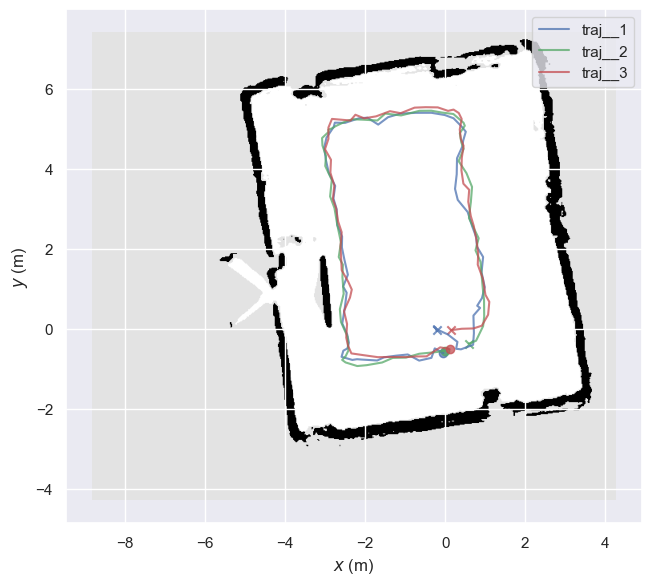}
        \caption{}
        \label{fig:exp-map-a}
    \end{subfigure}
    \hfill
    \begin{subfigure}[b]{0.45\textwidth}
        \includegraphics[height=7cm, width=8.5cm]{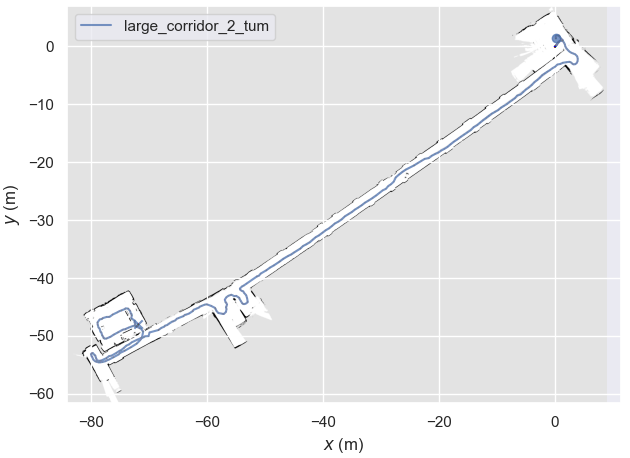}
        \caption{}
        \label{fig:exp-map-b}
    \end{subfigure}
    \caption{(a) Map of a corridor with three trajectories... (b) Map of a larger corridor...}
    \label{fig:experiment-mapping}
\end{figure}

\begin{figure}[htbp]
    \centering
    \includegraphics[width=\columnwidth]{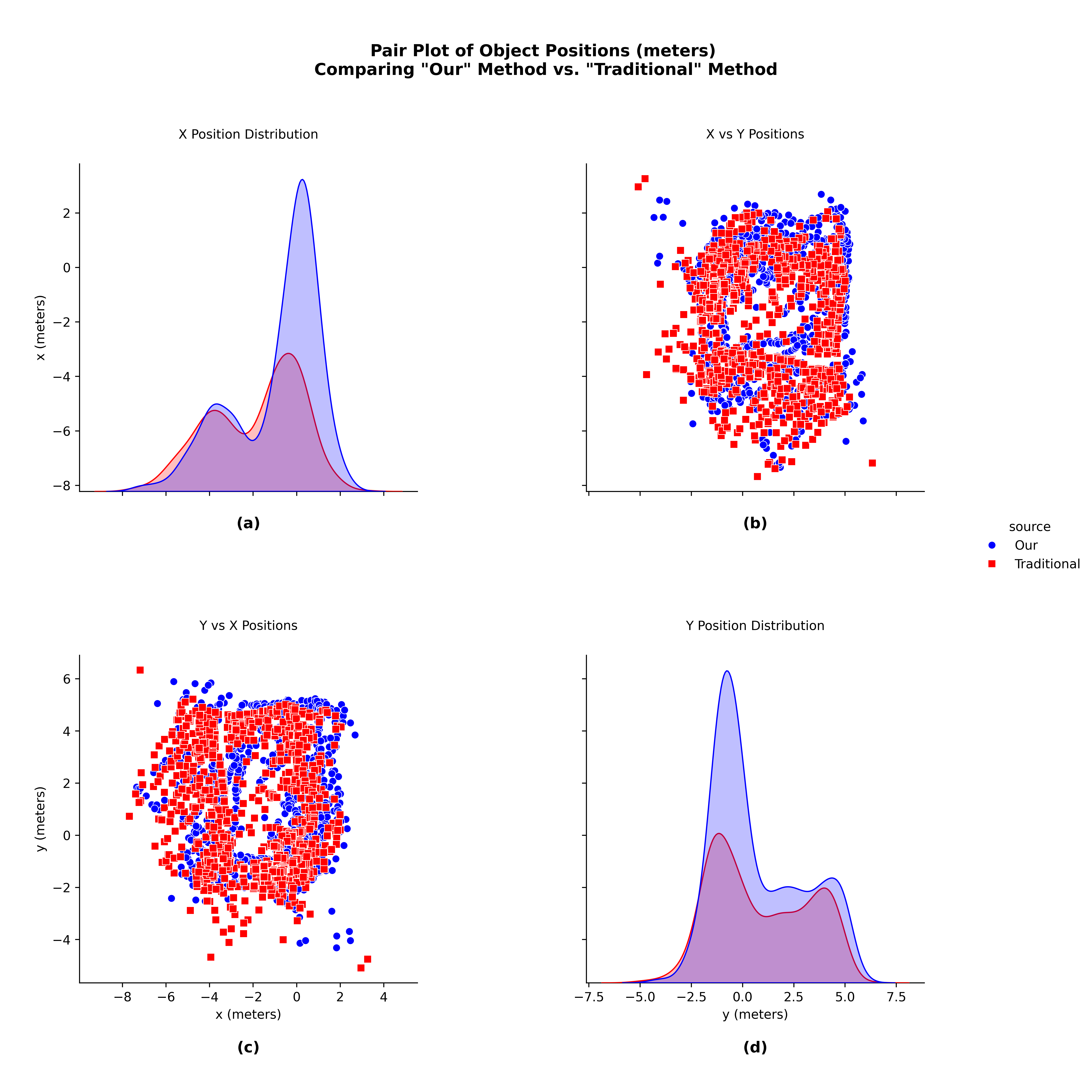}  
    \caption{Localization performance comparison between the proposed CNN-LiDAR-SLAM and \cite{bdcc7010043}.(a) X position distribution.(b) X vs. Y positions. (c) Y vs. X positions. (d) Y position distribution. The structured spatial alignment in our method minimizes localization drift and improves trajectory consistency.}
    \label{fig:Comparison-of-localization}
\end{figure}

In the 100- and 90-meter corridors, the proposed system achieved ATE values of 0.25 meter and 0.3 meter, respectively, as shown in Fig.~\ref{fig:experiment-mapping}~(b). This high localization accuracy was maintained without the use of loop closure techniques, which are commonly used in SLAM to mitigate drift over long distances. The proposed system exhibited minimal drift throughout the test trajectory, despite the absence of loop closure. This stability can be attributed to the effective integration of LiDAR data with IMU measurements, which supported continuous localization in environments with repetitive structures, as shown in Fig. \ref{fig:traj_comparasion}. Real-time sensor fusion effectively minimized cumulative errors along longer trajectories, maintaining low error rates throughout. The system performed consistently in environments of different sizes and geometries, demonstrating suitability for dynamic and structurally repetitive settings, such as emergency response operations. Furthermore, Fig.~\ref{fig:Comparison-of-localization} provides a detailed comparison of object positioning accuracy between the proposed CNN-LiDAR-SLAM and the Traditional SLAM method. Accurate object positioning plays a critical role in maintaining high localization accuracy by ensuring structured and consistent spatial mapping. Subfigures (a) and (d) illustrate the x and y position distributions, where our method demonstrates a more concentrated spread, reducing positional drift. The x-y position scatter plots in (b) and (c) further highlight the differences in spatial consistency, where CNN-LiDAR-SLAM shows a more structured and aligned distribution compared to the more dispersed pattern observed in the Traditional SLAM approach \cite{bdcc7010043}. These improvements directly contribute to reducing cumulative localization errors, complementing the results in Fig. ~\ref{fig:Comparison-of-localization} and reinforcing the observed ATE values in Fig. ~\ref{fig:experiment-mapping}~(b).

Table \ref{tab:comparison-key-Metrics} provides a quantitative performance comparison between CNN-LiDAR-SLAM and the baseline method of \cite{bdcc7010043}. For indoor obstacles, CNN-LiDAR-SLAM achieved a mean localization error of 1.02 m, outperforming the mean error of the baseline method of 1.38 m, which represents a 26.09\% improvement in localization accuracy. This increased accuracy is attributed to the introduction of the object detection integration method in the SLAM, which enables more precise differentiation between static and dynamic elements in complex environments. The results underscore CNN-LiDAR-SLAM robustness in scenarios with substantial environmental clutter and dynamic obstacles. The inclusion of object detection not only refines the accuracy of the mapping, but also improves the resilience of the system, ensuring stable performance in varying indoor settings.

\begin{table}[htbp]
    \centering
    \caption{\MakeUppercase{Performance comparison of key metrics between the proposed CNN-LiDAR-SLAM and the baseline method from \cite{bdcc7010043}}}
    \begin{tabular}{cccc}
    \hline
    Evaluation Metric  & CNN-LiDAR-SLAM & \cite{bdcc7010043} \\ 
    \hline
    $\mu_\epsilon$         & 1.02 meters              & 1.38 meters                 \\
    Improvement in Accuracy        & 26.09\%                  & -                           \\
    $\sigma_x$         & 1.20 meters              & 2.08 meters                 \\
    $\sigma_y$         & 2.04 meters              & 2.34 meters                 \\
    \hline
    \end{tabular}
    \label{tab:comparison-key-Metrics}
\end{table} 
\subsubsection{Object Detection Performance Evaluation}
To comprehensively validate the effectiveness of our CNN-based object detection module, we evaluated the module’s performance using standard metrics such as precision, recall, and F1-score. Additionally, we compared our approach with recent state-of-the-art detection methods from the literature, specifically those presented by Mochurad et al. \cite{bdcc7010043} and Fernandes et al. \cite{fernandes2021real}. These comparisons illustrate the superiority of our CNN method in terms of detection accuracy and reliability.

The detailed quantitative evaluation, shown in Table \ref{tab:cnn-comparison}, includes precision, recall, and F1 score.

\begin{table}[htbp]
    \centering
    \caption{\MakeUppercase{Performance comparison of key metrics between the proposed CNN-based detection and baseline methods from \cite{bdcc7010043,fernandes2021real}}}
    \begin{tabular}{cccc}
        \hline
        Method & Precision (\%) & Recall (\%) & F1-score \\
        \hline
        Proposed CNN Method & 92.5 & 91.3 & 0.919 \\
        Mochurad et al. \cite{bdcc7010043} & 86.3 & 83.5 & 0.849 \\
        Fernandes et al. \cite{fernandes2021real} & 88.9 & 86.5 & 0.877 \\
        \hline
    \end{tabular}
    \label{tab:cnn-comparison}
\end{table}

Our proposed CNN-based object detection demonstrates clear improvements with a precision of 92.5\%, recall of 91.3\%, and an F1-score of 0.919, reflecting significant advancements over the existing methods. Moreover, we have introduced a confusion matrix \ref{tab:cnn-comparison} to further clarify classification outcomes, distinguishing clearly among static structures, dynamic obstacles, and furniture categories.

\section{Discussion}
This study presents a 2D LiDAR and IMU-based localization system that improves accuracy, computational efficiency, and real-time adaptability. However, certain limitations remain. The system's reliance on 2D LiDAR makes it susceptible to performance degradation in featureless environments, such as long corridors and open spaces, where the lack of distinct landmarks can cause trajectory drift. Additionally, the experiments were limited in environmental diversity, notably lacking dynamic obstacles, varied lighting conditions, crowded spaces, and multi-floor structures. Furthermore, the absence of loop closure techniques can affect long-term mapping accuracy by causing cumulative drift over extended trajectories. The absence of a loop closure mechanism may lead to accumulated errors over long trajectories, affecting long-term mapping accuracy.

To overcome these challenges, future work should explore integrating additional sensors such as RGB-D cameras or 3D LiDAR to improve feature recognition and depth perception. This would enhance localization robustness in visually degraded settings. Furthermore, incorporating deep learning-based feature extraction could improve the system’s ability to differentiate static and dynamic elements, making it more effective in cluttered or dynamic environments. Additionally, optimizing computational efficiency through lightweight neural networks and edge computing techniques would make the system more suitable for real-time applications on mobile and robotic platforms.

These enhancements would significantly broaden the system’s applicability in robotics, emergency response, and industrial automation, providing a scalable and cost-effective solution for high-precision indoor localization. Future developments in multi-sensor fusion and AI-driven spatial awareness will further improve accuracy, robustness, and adaptability in complex and dynamic environments.

\section{Conclusion}

This study presented a portable 2D CNN-LiDAR-SLAM enhanced with an IMU sensor, designed to achieve high localization accuracy in indoor environments without the need for loop closure. Experimental results demonstrated that the proposed system consistently achieved a low ATE of 0.25 and 0.3 m in corridor environments, significantly reducing RMSE compared to established SLAM methods such as A-LOAM and SC-ALOAM. The integration of CNN-based object detection further improved the accuracy of mapping and localization, producing a performance gain of 26.09\% over comparable methods in cluttered environments. These findings highlight the potential of the system for reliable real-time localization in dynamic indoor settings. Future work will focus on optimizing memory usage and improving long-term performance, with plans to extend the system to 3D LiDAR for broader applicability and improved depth detection in complex environments.

\end{document}